\documentclass[10pt,twocolumn,letterpaper]{article}

\usepackage{iccv}
\usepackage{times}
\usepackage{epsfig}
\usepackage{graphicx}
\usepackage{amsmath}
\usepackage{amssymb}

\usepackage{multirow} 
\usepackage{pifont} 
\usepackage{slashbox}
\usepackage{diagbox}
\usepackage[dvipsnames]{xcolor} 
\usepackage{animategraphics} 
\usepackage{caption}  
\usepackage{kotex}  
\usepackage{mathtools}
\usepackage[symbol]{footmisc}
\renewcommand{\thefootnote}{\fnsymbol{footnote}}
\usepackage{textcomp}
\usepackage{graphicx}
\usepackage{booktabs}
\usepackage[toc,page]{appendix} 
\usepackage{color}
\definecolor{darkgreen}{rgb}{0,0.5,0.7}
\definecolor{purple}{rgb}{1,0,1}

\newif\ifdraft
\draftfalse

\ifdraft
    \newcommand{\jh}[1]  {\draft{purple}{#1}}
    \newcommand{\hj}[1]  {\draft{darkgreen}{#1}}
\else
    \newcommand{\jh}[1] {{\color{black}#1}}
    \newcommand{\hj}[1] {{\color{black}#1}}
\fi

\usepackage[pagebackref=true,breaklinks=true,bookmarks=false,colorlinks]{hyperref} 

\iccvfinalcopy 

\def\iccvPaperID{9691} 

\ificcvfinal\fi

\begin{document}

\title{XVFI: eXtreme Video Frame Interpolation}

\author{Hyeonjun Sim\footnotemark[1] \qquad\qquad Jihyong Oh\thanks{Both authors contributed equally to this work.} \qquad\qquad Munchurl Kim\thanks{Corresponding author.}\\
[0.5em]
Korea Advanced Institute of Science and Technology\\
{\tt\small \{flhy5836, jhoh94, mkimee\}@kaist.ac.kr}
}

\twocolumn[{%
\renewcommand\twocolumn[1][]{#1}%
\maketitle
\begin{center}\centering
    \setlength{\tabcolsep}{0.06cm}
    \setlength{\columnwidth}{2.81cm}
    \hspace*{-\tabcolsep}\begin{tabular}{cccccc}
            \animategraphics[width=\columnwidth, poster=3, autoplay, loop, final, nomouse, method=widget]{15}{VideoFigure_main/003TESTFast/}{00000}{00044}
        &
            \animategraphics[width=\columnwidth, poster=3, autoplay, loop, final, nomouse, method=widget]{15}{VideoFigure_main/004TESTMedium/}{00000}{00044}
        &
            \animategraphics[width=\columnwidth, poster=3, autoplay, loop, final, nomouse, method=widget]{15}{VideoFigure_main/045TESTMedium/}{00000}{00044}
        &
            \animategraphics[width=\columnwidth, poster=3, autoplay, loop, final, nomouse, method=widget]{15}{VideoFigure_main/078TESTFast/}{00000}{00044}
        &
            \animategraphics[width=\columnwidth, poster=3, autoplay, loop, final, nomouse, method=widget]{15}{VideoFigure_main/081TESTFast/}{00000}{00044}
        &
            \animategraphics[width=\columnwidth, poster=3, autoplay, loop, final, nomouse, method=widget]{15}{VideoFigure_main/146TESTMedium/}{00000}{00044}
        \\
            \footnotesize (a) \hj{96.6}
        &
            \footnotesize (b) \hj{71.0}
        &
            \footnotesize (c) \hj{34.9}
         &
            \footnotesize (d) \hj{40.6}
        &
            \footnotesize (e) \hj{196.5}
        &
            \footnotesize (f) \hj{152.2}
        \\
        \\
    \end{tabular}\vspace{-5mm}
	\captionof{figure}{Some examples of our \hj{X4K1000FPS} dataset, which contain diverse motions in 4K-resolution of 1000-fps. The numbers below the examples are the magnitude means of optical flows between two input frames in 30 fps. This is a video figure that can be best viewed with motion using Adobe\texttrademark\ Reader. It should be noted that they are rendered in down-scales at 15 fps for visualization convenience.}\vspace{-0.0cm}
	\label{fig:teaser}
\end{center}
}]

{
  \renewcommand{\thefootnote}%
    {\fnsymbol{footnote}}
  \footnotetext[1]{Both authors contributed equally to this work.}
  \footnotetext[2]{Corresponding author.}
}


\ificcvfinal\thispagestyle{empty}\fi

\begin{abstract}
   In this paper, we firstly present a dataset (X4K1000FPS) of 4K videos of 1000 fps with the extreme motion to the research community for video frame interpolation (VFI), and propose an extreme VFI network, called XVFI-Net, that first handles the VFI for 4K videos with large motion. The XVFI-Net is based on a recursive multi-scale shared structure that consists of two cascaded modules for bidirectional optical flow learning between two input frames (BiOF-I) and for bidirectional optical flow learning from target to input frames (BiOF-T). The optical flows are stably approximated by a complementary flow reversal (CFR) proposed in BiOF-T module. 
   During inference, the BiOF-I module can start at any scale of input while the BiOF-T module only operates at the original input scale so that the inference can be accelerated while maintaining highly accurate VFI performance. Extensive experimental results show that our XVFI-Net can successfully capture the essential information of objects with extremely large motions and complex textures while the state-of-the-art methods exhibit poor performance. \jh{Furthermore, our XVFI-Net framework also performs comparably on the previous lower resolution benchmark dataset, which shows a robustness of our algorithm as well}. All source codes, pre-trained models, and proposed X4K1000FPS datasets are publicly available at \url{https://github.com/JihyongOh/XVFI}.
\end{abstract}

\section{Introduction}

Video frame interpolation (VFI) converts low frame rate (LFR) contents to high frame rate (HFR) videos by synthesizing one or more intermediate frames between given two consecutive frames, and then the videos of fast motion can be smoothly rendered in an increased frame rate, thus yielding reduced motion judder \cite{mackin2015,kuroki2014effects,kuroki2007psychophysical,daly2001engineering}. Therefore, it is widely used for various practical applications, such as adaptive streaming \cite{wu2015modeling}, novel view interpolation synthesis \cite{flynn2016deepstereo}, frame rate up conversion \cite{meyer2015phase, castagno1996method, yu2019hierarchical}, slow motion generation \cite{jiang2018super,bao2019depth,niklaus2018context,niklaus2017video,liu2017video,peleg2019net} and video restoration \cite{kim2020fisr, wang2019edvr, haris2020space, tian2020tdan}. However, VFI is significantly challenging, which is attributed to diverse factors such as occlusions, large motions and change of light. Recent deep-learning-based VFI has been actively studied, showing remarkable performances \cite{xu2019quadratic,bao2019depth,choi2020channel,shen2020blurry,lee2020adacof,gui2020featureflow,niklaus2020softmax,zhang2019flexible,chi2020all,park2020bmbc}. However, they are often optimized for existing LFR benchmark datasets of low resolution (LR), which may lead to poor VFI performance, especially for videos of 4K resolution (4096$\times$2160) or higher with very large motion \cite{ahn2019fasta, kim2020fisr}. Such 4K videos often contain frames of fast motion with extremely large pixel displacements for which conventional convolutional neural networks (CNNs) do not effectively work with receptive fields of limited sizes.

\begin{figure} [ht]
\centering
\includegraphics[width=\columnwidth]{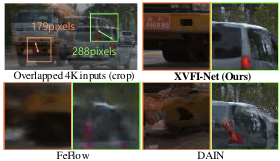}
\caption{VFI results for extreme motions. Our XVFI-Net can generate a more stable intermediate frame with very large motions than two recent SOTA methods, FeFlow \cite{gui2020featureflow} and DAIN \cite{bao2019depth}, which are newly trained on our dataset for fair comparisons.}
\label{fig: XVFI_sota_comparison(teaser)}
\end{figure}

To solve the above issues for deep learning-based VFI methods, we directly photographed 4K videos to construct a high-quality HFR dataset of high resolution, called X4K1000FPS. Fig. \ref{fig:teaser} shows some examples of our X4K1000FPS dataset. As shown, our videos of 4K resolution have extremely large motions and occlusions.

We also first propose an extreme VFI model, called XVFI-Net, that is effectively designed to handle such a challenging dataset of 4K@1000fps. Instead of directly capturing extreme motions through consecutive feature spaces with deformable convolution as recent trends in video restoration \cite{gui2020featureflow, xiang2020zooming, wang2019edvr,tian2020tdan,kang2020deep}, or using very large-sized pretrained networks with extra information such as contexts, depths, flows and edges \cite{bao2019depth,zhang2019flexible, niklaus2018context, gui2020featureflow}, our XVFI-Net is simple but effective, which is based on a recursive multi-scale shared structure. The XVFI-Net has two cascaded modules: one for the bidirectional optical flow learning between two input frames (BiOF-I) and the other for the bidirectional optical flow estimating from target to the inputs (BiOF-T). The BiOF-I and BiOF-T modules are trained in combination with multi-scale losses. However, once trained, the BiOF-I module can start from any down-scaled input upward while the BiOF-T module only operates at the original input scale at inference, which is computationally efficient and helps to generate an intermediate frame at any target time instance. Structurally, the XVFI-Net is adjustable in terms of the number of scales for inference according to the input resolutions or the motion magnitudes, even if training is once over. We also propose a novel optical flow estimation from time $t$ to those of the inputs, called a complementary flow reversal (CFR) that effectively fills the holes by taking complementary flows. Extensive experiments are conducted for fair comparison and our XVFI-Net that has a relatively smaller complexity outperforms previous VFI SOTA methods on our X4K1000FPS, especially for extreme motions as shown in Fig. \ref{fig: XVFI_sota_comparison(teaser)}. \hj{A further experiment on the previous LR-LFR benchmark dataset also demonstrates the robustness of our XVFI-Net.} Our contributions are summarized as:
\vspace{-1.5mm}
\begin{itemize}
\item We \hj{first} propose a high-quality of HFR video dataset of 4K resolution, called X4K1000FPS (4K@1000fps) which contains a wide variety of  textures, extremely large motions, zoomings and occlusions.

\item We propose the CFR that can generate stable optical flow estimation from time $t$ to the input frames, boosting both qualitative and quantitative performances.

\item Our proposed XVFI-Net can start from any down-scaled input upward, which is adjustable in terms of the number of scales for inference according to the input resolutions or the motion magnitudes.

\item Our XVFI-Net achieves state-of-the-art performance on the testset of X4K1000FPS with a significant margin compared to the previous VFI SOTA methods while having computational efficiency with a small number of filter parameters. All source codes and proposed X4K1000FPS dataset are publicly available at \url{https://github.com/JihyongOh/XVFI}.
\end{itemize}



\section{Related Work}
\subsection{Video Frame Interpolation}
Most VFI methods can be categorized into \hj{optical} flow- or kernel-based \cite{liu2017video, niklaus2017video,jiang2018super, niklaus2018context,peleg2019net,xu2019quadratic,bao2019depth,ahn2019fasta,ahn2019fastb,lee2020adacof,park2020bmbc, niklaus2020softmax} and pixel hallucination-based \cite{gui2020featureflow, xiang2020zooming, choi2020channel, shen2020blurry, kim2020fisr} methods. 
\noindent
\textbf{Flow-based VFI}. Super-SloMo \cite{jiang2018super} \jh{first} linearly combines predicted optical flows between two input frames to approximate flows from the target intermediate frame to the input frames. Quadratic video frame interpolation \cite{xu2019quadratic} utilizes four input frames to cope with nonlinear motion modeling by quadratic approximation, which limits the VFI generalization when two input frames are given. It also proposes flow reversal (projection) for more accurate image warping. On the other hand, DAIN \cite{bao2019depth} gives different weights of overlapped flow vectors depending on the object depth of the scene via a flow projection layer. 
However, DAIN employs and fine-tunes both PWC-Net \cite{sun2018pwc} and MegaDepth \cite{li2018megadepth}, which is computationally burdened for inferring intermediate HR frames. AdaCoF proposes a generalized warping module to deal with complex motion \cite{lee2020adacof}. However, it is not adaptive to handle the frames of higher resolutions \jh{due to a fixed dilation degree}, after once trained.

\noindent
\textbf{Pixel Hallucination-based VFI}. FeFlow \cite{gui2020featureflow} has benefited from deformable convolution \cite{dai2017deformable} to the center frame generator by replacing optical flows with offset vectors. Zooming Slow-Mo \cite{xiang2020zooming} also interpolates middle frames with the help of deformable convolution in the feature domain. However, since these methods directly hallucinate pixels unlike the flow-based VFI methods, the predicted frames tend to be blurry when fast-moving objects are present. 

Most importantly, the aforementioned VFI methods are difficult to operate on the entire HR frames \textit{at once}, due to their heavy computational complexity. On the other hand, our XVFI-Net is designed to efficiently operate on the entire 4K frame input at once with a smaller number of parameters and is capable of effectively capturing large motions.

\subsection{Networks for Large Pixel Displacements}
PWC-Net \cite{sun2018pwc} is a state-of-the-art optical flow estimator that \hj{has been} adopted in several VFI methods for pretrained flow estimators \cite{xu2019quadratic, bao2019depth, niklaus2020softmax}. Since PWC-Net has a 6-level feature pyramid structure to have larger sizes of receptive fields, it enables to effectively predict large motions. IM-Net \cite{peleg2019net} also adopts a multi-scale structure to cover large displacements of objects in adjacent frames while the coverage is limited in the size of the adaptive filters. Despite of the multi-scale pyramid structures, the above methods lack adaptivity because the coarsest level of each network is fixed after once trained, i.e. each scale level \hj{consists of} its own (\textit{not shared}) parameters. The RRPN \cite{zhang2019flexible} shares weight parameters across different scale levels in a flexible recurrent pyramid structure. However, it only infers the centered frame, not at arbitrary time instances. So it can only synthesize recursively the intermediate frames of time at a power of 2. As a result, the prediction errors are accumulated as intermediate frames are generated iteratively between the two input frames. Therefore, RRPN has limited temporal flexibility for VFI at arbitrary target time instance \textit{t}.

Distinguished from the above methods, our proposed XVFI-Net has a scalable structure with shared parameters for various input resolutions. Different from RRPN \cite{zhang2019flexible}, the XVFI-Net is structurally divided into the BiOF-I and BiOF-T modules, which allows predicting an intermediate frame at arbitrary time $t$ \hj{with the help of the complementary flow reversal} in an efficient way. That is, the BiOF-T module can be skipped at the down-scaled levels in inference so that our model can infer the intermediate frame of 4K at once, without any patch-wise iteration unlike all other previous methods, which \jh{can} be applied in real-world applications.

\section{Proposed X4K1000FPS Dataset}

Although numerous methods for VFI have been both trained and evaluated over the diverse benchmark datasets, such as Adobe240fps \cite{su2017deep}, DAVIS \cite{perazzi2016benchmark}, UCF101 \cite{soomro2012ucf101}, Middlebury \cite{baker2011database} and Vimeo90K \cite{xue2019video}, none of the datasets contains rich amounts of 4K videos with HFR. These limits the study of elaborate VFI methods required for VFI applications for targeting very high resolution videos.

To tackle the challenging extreme VFI task, we provide a rich set of 4K@1000fps video that we photographed using a Phantom Flex4K\texttrademark\ camera with the 4K spatial resolution of 4096$\times$2160 at 1,000 fps, producing 175 video scenes, each with 5,000 frames by shooting for 5 seconds. 

\hj{In order to select valuable data samples for VFI, we estimated bidirectional occlusion maps and optical flows of every 32 frames of the scenes using IRR-PWC \cite{hur2019iterative}. The occlusion map indicates part of the objects to be occluded in the next frames. The occlusion makes optical flow estimation and frame interpolation challenging \cite{wang2018occlusion, bao2019depth, hur2019iterative}. Thus, we manually selected 15 scenes as our testset, called X-TEST, by considering the degrees of occlusion, optical flow magnitudes and scene diversity.} 
Each scene for X-TEST simply contains one test sample that consists of two input frames in a temporal distance of 32 frames and approximately corresponds to 30 fps. The test evaluation is set to interpolate \hj{7} intermediate frames, which results in the consecutive frames of 240 fps.
\hj{For the training dataset, called X-TRAIN, we cropped and selected 4,408 clips of 768$\times$768-sized and the lengths of 65 consecutive frames by considering the amounts of occlusion.} More details are described in \textit{Supplementary Material}.

\hj{Table \ref{dataset_statistics} compares the statistics of datasets: Vimeo90K \cite{xue2019video},  Adobe240fps \cite{su2017deep}, our X-TEST and X-TRAIN. We estimated the occlusion range in [0,255] and optical flow magnitudes \cite{hur2019iterative} between input pairs and calculated their percentiles for each dataset. As shown in Table \ref{dataset_statistics}, our datasets contain comparable occlusion but significantly larger motion, compared to the previous VFI datasets.} 

\begin{table}
\begin{center}
\scalebox{0.8}{
\begin{tabular}{ c||c|c|c||c|c|c }
\hline
\multirow{2}{*}{Dataset} & \multicolumn{3}{c||}{Occlusion \cite{hur2019iterative}} & \multicolumn{3}{c}{Flow magnitude \cite{hur2019iterative}} \\ 
 & 25$_{th}$ & 50$_{th}$ & 75$_{th}$ & 25$_{th}$ & 50$_{th}$ & 75$_{th}$  \\ \hline  
Vimeo90K \cite{xue2019video} & \underline{6.8} & \textbf{11.9} & \textbf{18.1} & 3.1 & 4.9 & 7.1 \\ \hline
Adobe240fps \cite{su2017deep} & 0.8 & 1.7 & 3.2 & 3.8 & 8.9 & 16.3 \\ \hline
X-TEST (ours) & 2.1 & 5.6 & \underline{17.7} & \textbf{23.9} & \textbf{81.9} & \textbf{138.5} \\ \hline
X-TRAIN (ours) & \textbf{6.9} & \underline{10.1} & 15.7 & \underline{5.5} & \underline{18.0} & \underline{59.5} \\ \hline
\multicolumn{7}{l}{25$_{th}$, 50$_{th}$ and 75$_{th}$ represent percentiles of each datset.}\\
\end{tabular}}\vspace{-3mm}
\end{center}
\caption{The occlusion and optical flow magnitude statistics of VFI datasets: 3,782 test triplets of Vimeo90K \cite{xue2019video}, randomly selected 200 clips of Adobe240fps \cite{su2017deep}, 15 clips of X-TEST and 4,408 clips of X-TRAIN.}
\label{dataset_statistics}
\end{table}

\section{Proposed Method : XVFI-Net Framework}
\subsection{Design Considerations}
Our XVFI-Net aims at interpolating an intermediate frame $I_t$ at an arbitrary time $t$ between two consecutive input frames, $I_0$ and $I_1$, of HR with extreme motion. 

\noindent
\textbf{Scale Adaptivity}.
An architecture with a fixed number of scale levels like PWC-Net \cite{sun2018pwc} is difficult to adapt to various spatial resolutions of the input video, because the structure in each scale level is \textit{not shared} across different scale levels, so the new architecture with an increased scale depth needs to be retrained. In order to have a scale adaptivity to various spatial resolutions of input frames, our XVFI-Net is designed to have optical flow estimation starting at any desired coarse scale level, adapting to the degree of motion magnitudes in the input frames. To do so, our XVFI-Net shares its parameters across different scale levels.

\begin{figure}
\vspace*{-0.1in}
\includegraphics[width=0.48\textwidth]{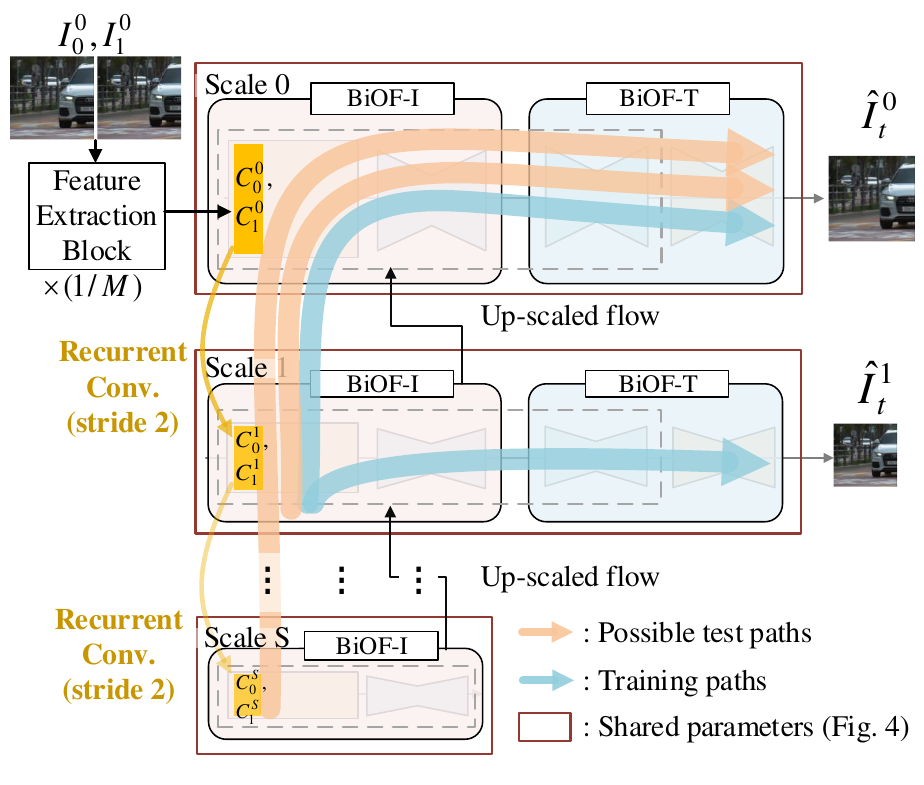}
\caption{Adjustable and efficient scalability of our XVFI-Net framework. Even if the lowest scale depth $S_{trn}$ during training is set to 1 in this example, inference can start from any scale level.}
\label{fig: framework}
\vspace*{-0.1in}
\end{figure}

\noindent
\textbf{Capturing Large Motion}.
In order to effectively capture a large motion between two input frames, \jh{the Feature Extraction Block of XVFI-Net first reduces the spatial resolution of two input frames by a module scale factor $M$ via a strided convolution, thus yielding the spatially reduced feature that is then converted to two contextual feature maps $C_0^{0}$ and $C_1^{0}$. The Feature Extraction Block in Fig. \ref{fig: framework} is composed of the strided convolution and two residual blocks \cite{he2016deep}. Then, XVFI-Net at each scale level estimates optical flows from target frame $I_t$ to two input frames in the reduced spatial size by $M$. The predicted flows are upscaled ($\times M$) to warp the input frames at each scale level to time $t$.}

\subsection{XVFI-Net Architecture} 
\begin{figure*}
\centering
\includegraphics[width=\textwidth]{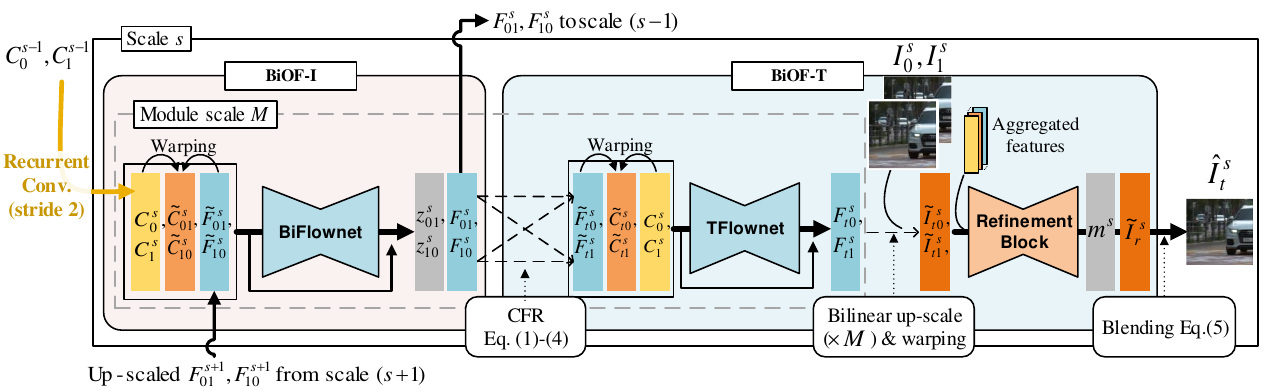}
\caption{The architecture of our proposed XVFI-Net in scale $s$. }
\label{fig: architecture}
\end{figure*}

\noindent
\textbf{BiOF-I module}. Fig. \ref{fig: architecture} shows our XVFI-Net architecture in scale \textit{s}, \jh{where $I^s$ denotes bicubicly down-scaled by $1/2^s$. First, contextual pyramid $\textbf{C}=\{C^{s}\}$ is recurrently extracted from $C^0_0$ and $C^0_1$ via a stride 2 convolution, and then utilized as an input for XVFI-Net at each scale level $s$ $(s=0,1,2,...)$, where $s=0$ denotes the scale of the original input frames.} Let $F^{s}_{t_at_b}$ denotes optical flow from time $t_a$ to $t_b$ at scale $s$. $F^{s}_{01}$ and $F^{s}_{10}$ are the bidirectional flows between input frames at scale $s$. $F^{s}_{t0}$ and $F^{s}_{t1}$ are the bidirectional flows from $I^s_t$ to $I^s_0$ and $I^s_1$, respectively. 

The estimated flows $F^{s+1}_{01}, F^{s+1}_{10}$ from the previous scale ($s+1$) are $\times 2$ bilinearly up-scaled to be set as the initial flows for the current scale $s$, i.e., $\Tilde{F}^s_{01}=F^{s+1}_{01}\uparrow_2, \Tilde{F}^s_{10}=F^{s+1}_{10}\uparrow_2$. To update the initial flows in the current scale, $C^s_0$ and $C^s_1$ are first warped by the initial flows, that is, $\Tilde{C}^s_{01}=W(\Tilde{F}^s_{01}, C^s_1)$ and $\Tilde{C}^s_{10} =W(\Tilde{F}^s_{10}, C^s_0)$, respectively, where $W$ is a backward warping operation \cite{jaderberg2015spatial}. Then $\Tilde{C}^s_{01}, \Tilde{C}^s_{10}, C^s_0, C^s_1 $ together with $\Tilde{F}^s_{01},\Tilde{F}^s_{10}$ are fed to an auto-encoder-based BiFlownet as in Fig. \ref{fig: architecture} to output residual flows over the initial flows and a trainable importance mask $z$ \cite{niklaus2020softmax}. Then $F^{s}_{01}, F^{s}_{10}$ are obtained. They are then fed as input to the BiOF-T module and are also used as the initial flows to the next scale $s-1$.

\noindent    
\textbf{BiOF-T module}. Hereafter, we omit superscript $s$ for the notion of feature tensors at each scale, unless mentioned. Although the linear approximation with optical flows $F_{01}, F_{10}$ \cite{jiang2018super} or the flow reversal of $F_{0t}, F_{1t}$ \cite{xu2019quadratic} allows to estimate the flows $F_{t0}, F_{t1}$ at arbitrary time $t$, there are few shortcomings. The linear approximation is inaccurate to predict $F_{t0}$ and $F_{t1}$ for fast-moving objects because the anchor points of $F_{01}$ and $F_{10}$ are severely misaligned. On the other hand, the flow reversal can align the anchor points but \textit{holes} may appear in estimated $F_{t0}$ and $F_{t1}$. To stabilize the performance of the flow reversal, we take complementary advantages of both the linear approximation and flow reversal. So, a stable optical flow estimate from time $t$ to 0 or 1 can be computed by a normalized linear combination of a negative anchor flow and a complementary flow, which we call a complementary flow reversal (CFR). The resulting complementary reversed optical flow maps, ${\Tilde{F}_{t0}}$ and ${\Tilde{F}_{t1}}$, from time $t$ to 0 and 1 are given by, 
\vspace{-1mm}
\begin{flalign}
    & {\Tilde{F}_{t0}^{\mathbf{x}}}=
    \cfrac{ (1-t)\sum_{\mathcal{N}_{0}} w_0 \! \cdot \! (-{F_{0t}^{\mathbf{y}}}) + t\sum_{\mathcal{N}_{1}} \! w_1 \! \cdot {F_{1\cdot(1\text{-}t)}^{\mathbf{y}}}}
    { (1-t)\sum_{\mathcal{N}_{0}} \! w_0 \! + t\sum_{\mathcal{N}_{1}} w_1 }
    \label{eq_fwarp1}\\
    & {\Tilde{F}_{t1}^{\mathbf{x}}}=
    \cfrac{ (1-t)\sum_{\mathcal{N}_{0}} \! w_0 \! \cdot {F_{0\cdot(1\text{-}t)}^{\mathbf{y}}} + t\sum_{\mathcal{N}_{1}} \! w_1 \! \cdot (-{F_{1t}^{\mathbf{y}}}) }
    { (1-t)\sum_{\mathcal{N}_{0}} w_0 + t\sum_{\mathcal{N}_{1}} w_1 }
    \label{eq_fwarp2}
\end{flalign}
where \textbf{x} denotes a pixel location at time $t$ and \textbf{y} is at time 0 or 1. $w_{i} = z^{\mathbf{y}}_{i}\cdot G(|\mathbf{x}-(\mathbf{y}+{F_{it}^{\mathbf{y}}})|)$ is a Gaussian weight depending on the distance between \textbf{x} at time $t$ and $\mathbf{y}+{F_{it}^{\mathbf{y}}}$ at time $i$ (= 0 or 1) while also considering the learnable importance mask of each flow by $z^{\mathbf{y}}_i$ \cite{niklaus2020softmax}. Also, $-{F_{0t}^{\mathbf{y}}}$ (or $-{F_{1t}^{\mathbf{y}}}$) and  ${F_{1\cdot(1\text{-}t)}^{\mathbf{y}}}$ (or ${F_{0\cdot(1\text{-}t)}^{\mathbf{y}}}$) in Eq. \ref{eq_fwarp1} (or Eq. \ref{eq_fwarp2}) are defined as a negative anchor flow and a complementary flow, respectively. Furthermore, the anchor flows are normalized flows that can be calculated as $F_{0t}=t F_{01}$ and $F_{1t}=(1-t)F_{10}$ to intermediate time $t$. 
It should be noted in Eq. \ref{eq_fwarp1} and Eq. \ref{eq_fwarp2} that the complementary flows are also \textit{normalized} as $F_{1\cdot(1\text{-}t)}=t F_{10}$ and $F_{0\cdot(1\text{-}t)}=(1-t) F_{01}$ which complementally fill the holes occurred in the reversed flows. By doing so, we can fully exploit the temporal-densely captured X4K1000FPS dataset to train our XVFI-Net for VFI at arbitrary time \textit{t}. The neighborhoods of \textbf{x} are defined as,
\vspace{-0.5mm}
\begin{align}
    & \mathcal{N}_{0}=\{{\mathbf{y}} \ | \ \text{round}({\mathbf{y}}+{F_{0t}^\mathbf{y}})={\mathbf{x}}\}
    \label{eq_neighbor1}\\
    & \mathcal{N}_{1}=\{{\mathbf{y}} \ | \ \text{round}({\mathbf{y}}+{F_{1t}^\mathbf{y}})={\mathbf{x}}\}.
    \label{eq_neighbor2}
\end{align}
\vspace{-0.5mm}
To refine the bidirectional flow approximates $\Tilde{F}_{t0},\Tilde{F}_{t1}$, we rewarp the feature maps ($C_0, C_1$) to $\Tilde{C}_{t0}$ and $\Tilde{C}_{t1}$ by $\Tilde{F}_{t0}$ and $\Tilde{F}_{t1}$, respectively. We concatenate and feed $C_0, C_1,\Tilde{C}_{t0},\Tilde{C}_{t1}$, and $\Tilde{F}_{t0},\Tilde{F}_{t1}$ to the auto-encoder-based TFlownet as in Fig. \ref{fig: architecture} (similarly to refine $\Tilde{F}_{01},\Tilde{F}_{10}$). The outputs of TFlownet are used to compose refined flows $F_{t0},F_{t1}$ which are then bilinearly up-scaled ($\times M$) back to the size of $I^s_t$. The flow estimation in the spatially reduced size by $M$ has three advantages: (i) enlarged receptive fields, (ii) lowered computational costs and (iii) smooth optical flows. This strategy maximizes the benefit of flow-based VFI that can fully utilize the texture information of the original input frames by warping them with the estimated flows, compared to the hallucination-based methods that suffer from a lack of sharpness in restoration from down-scaled feature maps. \hj{The above up-scaled flows are used to warp the input frames $I^s_{0}$ and $I^s_{1}$ to be $\Tilde{I}^s_{t0}$ and $\Tilde{I}^s_{t1}$, respectively. The $C^s_0, C^s_1, \Tilde{C}^s_{t0}, \Tilde{C}^s_{t1}, F^s_{t0}, F^s_{t1}, I^s_{0}, I^s_{1}, \Tilde{I}^s_{t0}$ and $\Tilde{I}^s_{t1}$ are all aggregated to be fed into the U-Net \cite{ronneberger2015u}-based Refinement Block. Then, both the generated occlusion mask $m^s$ and residual image $\Tilde{I}^s_{r}$ are finally used to blend the warped frames $\Tilde{I}^s_{t0}$ and $\Tilde{I}^s_{t1}$}, which is given by, 
\begin{flalign}
    \hat{I^s_t} = 
    \frac{(1-t)\!\cdot\!m^s\!\cdot{\Tilde{I}^s_{t0}} + t\!\cdot\!(1-m^s)\!\cdot\! {\Tilde{I}^s_{t1}}}{(1-t) \cdot m^s + t \cdot (1-m^s)} + \Tilde{I}^s_{r}
    \label{eq_bleding}
\end{flalign}
where $\hat{I^s_t}$ is the final result of each scale level $s$.

\subsection{Adjustable and Efficient Scalability}
\noindent
\textbf{Adjustable Scalability}. \hj{Fig. \ref{fig: framework} shows a VFI framework of our XVFI-Net} that can begin from any scale level \jh{by $\times 1/2^s$ \hj{recurrent} down-scaling the contextual feature map $C^0_0$ and $C^0_1$}, and predicts the coarsest optical flow to capture extreme motion effectively. Then the estimated flows $F^{s}_{01}$, $F^{s}_{10}$ are transmitted to the next scale $s-1$, and the flow is updated gradually to the original scale $s = 0$. We aim that the number of scales can be decided for inference, adaptive to the spatial resolution and degree of motion magnitudes for the input frames, even after once trained. To generalize the XVFI-Net learning for the input of any scale level, a multi-scale reconstruction loss in Eq. \ref{eq_rec_loss} is applied for every output $\hat{I^s_t}$ for the selected scale depth $S_{trn}$ during training. 

\noindent
\textbf{Efficient Scalability}. As shown in Fig. \ref{fig: framework}, the computation through the BiOF-T module is always taken place at the original scale ($s=0$) during inference no matter which scale level the BiOF-I starts from, which are denoted as the arrows in the light orange color. Since $F^s_{01}$ and $F^s_{10}$ are the only information that passes across different scale levels through the BiOF-I module (from the previous scale to the next scale level) as shown in Fig. \ref{fig: framework}, we only pass the two optical flows recursively until reaching the original scale level. Then, the BiOF-T module processes $F^{s=0}_{10}$ and $F^{s=0}_{01}$ to estimate $F^{s=0}_{t1}$ and $F^{s=0}_{t0}$ only at the original scale level. This is architecturally very beneficial because (i) the BiOF-I module is responsible to stably capture extreme motion by recursively learning the bidirectional flows between input time instances 0 and 1 across multiple scale levels, and (ii) the BiOF-T module finely predicts the bidirectional flows in the original scale only from any target time $t$ to times 0 and 1 based on the stably estimated flows $F^{s=0}_{10}$ and $F^{s=0}_{01}$, unlike the RRPN \cite{zhang2019flexible}.

\noindent
\textbf{Loss Functions}.
We adopt a multi-scale reconstruction loss to train the shared parameters of our XVFI-Net. To further encourage the smoothness of the obtained optical flow, the first-order edge-aware smoothness loss is used for $F^0_{t0}$ and $F^0_{t1}$ at the original scale \cite{jonschkowski2020matters}. The total loss function is a weighted sum of the two loss functions as follows: 
\begin{align}
    & \mathcal{L}_{total} = \mathcal{L}_{r} + \lambda_s \cdot \mathcal{L}_{s}
    \label{eq_total_loss} \\
    & \mathcal{L}_{r} = \textstyle\sum_{s=0}^{S_{trn}} \lVert \hat{I}_{t}^{s}-I_{t}^{s}\rVert_{1}
    \label{eq_rec_loss}\\
    & \mathcal{L}_{s} = \textstyle\sum_{i=0,1} \text{exp}(-{e^2} \sum_{c} \left\lvert {\nabla_{\mathbf{x}} I^0_{t_c}} \right\rvert)^\intercal \cdot \left\lvert {\nabla_{\mathbf{x}} F^0_{ti}} \right\rvert 
    \label{eq_smooth_loss}
\end{align}
where $c,e^2$ and $\mathbf{x}$ denote color channel index, an edge weighting factor and a spatial coordinate, respectively.

\section{Experiment Results}
The proposed X-TRAIN dataset contains \hj{4,408 clips of the sizes of 768$\times$768 and the lengths of 65 consecutive frames}. Each training sample is randomly fetched on the fly from each clip. A training sample is defined as a triplet with two input frames ($I_{0}, I_{1}$) and one target frame ($I_{t}, 0 < t < 1$). The temporal distance between $I_{0}$ and $I_{1}$ is randomly selected in the range [2, 32] where $I_{t}$ is also randomly determined between the selected $I_{0}$ and $I_{1}$. By doing so, our training samples are stochastically well obtained by fully exploiting our X-TRAIN dataset of temporally dense video clips to learn various $t$ accordingly.

The weights of the XVFI-Net are initialized with Xavier initialization \cite{glorot2010understanding} and the mini-batch size is set to 8. XVFI-Net is trained via total of 110,200 iterations (200 epochs) by using the Adam optimizer \cite{kingma2014adam} with the initial learning rate of $10^{-4}$, reduced by a factor of 4 at [100, 150, 180]-$th$ epoch. The hyperparameter $M$, $\lambda_s$ and $e$ are set to 4, 0.5 and 150, respectively. We also randomly crop $384\times384$-sized patches from the original size of X-TRAIN and randomly flip both spatial and temporal directions for data augmentation. Training takes about a half-day with an NVIDIA TITAN RTX\texttrademark\ GPU with PyTorch.

\subsection{Comparison to the Previous Methods}
We compare our XVFI-Net with three previous VFI methods, DAIN \cite{bao2019depth}, FeFlow \cite{gui2020featureflow} and AdaCoF \cite{lee2020adacof}, whose training codes are \textit{publicly available}. DAIN can generate the interpolated frame at arbitrary time \textit{t} at once and the latter two can only synthesize the intermediate frame at the power of 2 in an iterative manner \hj{during the inference}. 

For a fair comparison, we \textit{retrain} the three previous methods on X-TRAIN under their original hyperparameter settings except the patch size of 384$\times$384 and the total iterations of 110,200. For further comparison, we also use the original pretrained models of the three methods, which are denoted as the subscript \textit{o} to distinguish from their retrained models with the subscript \textit{f} on X-TRAIN. \jh{The lowest scale depths for our XVFI-Net are set to 3 for training ($S_{trn}$) and 3 or 5 for testing ($S_{tst}$)}. \hj{We evaluate their performances for 7 interpolated frames per scene (multi-frame interpolation $\times8$) on X-TEST in terms of three evaluation metrics: PSNR, SSIM \cite{wang2004image} and tOF \cite{Chu2020TecoGAN} that measures the \textit{temporal} consistency for the pixel-wise difference of motions (the lower, the better).}
\jh{We also evaluate each method for 7 interpolated frames per clip on the Adobe240fps dataset \cite{su2017deep}, where 200 \hj{nonuplets} clips are randomly extracted with $1280\times720$ (HD) at 240fps.}

\begin{figure}
\includegraphics[scale=0.7]{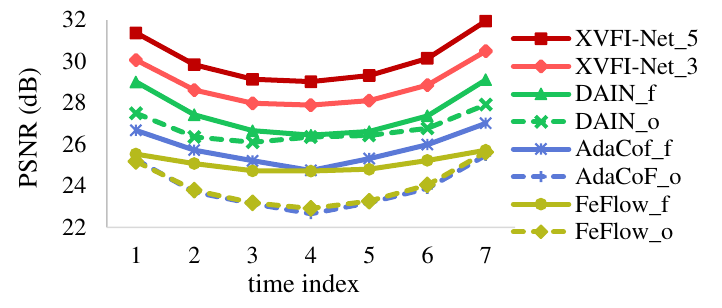}
\caption{PSNR profiles for multi-frame interpolation results ($\times8$) on X-TEST.}
\label{fig: multi-frame interplation}
\end{figure}

\begin{table}
\begin{center}
\scalebox{0.74}{
\begin{tabular}{ c||c||c||c|c }
\hline
\multirow{2}{*}{Methods ($\times$N)} & X-TEST & Adobe240fps & \multirow{2}{*}{\#P $\downarrow$} & \multirow{2}{*}{R$_{t}$ $\downarrow$}  \\
 & (PSNR/SSIM/tOF) & (PSNR/SSIM) & & \\
\hline
AdaCoF$_{o}$ ($\times$5.8) & 23.90/0.727/6.89 & 25.26/0.785 & \textcolor{blue}{\underline{21.8}} & \textcolor{red}{\textbf{0.005}} \\ 
AdaCoF$_{f}$ \cite{lee2020adacof} & 25.81/0.772/6.42 & 25.21/0.791 & \textcolor{blue}{\underline{21.8}} & \textcolor{red}{\textbf{0.005}}\\ 
\hline
FeFlow$_{o}$ ($\times$5.3) & 24.00/0.756/6.59 & 25.18/0.785 & 102.5 & \hj{1.681} \\
FeFlow$_{f}$ \cite{gui2020featureflow} & 25.16/0.783/6.54 & 24.17/0.780 & 102.5 & \hj{1.681} \\
\hline
DAIN$_{o}$ ($\times$9.3) & 26.78/0.807/3.83 & 29.89/\textcolor{blue}{\underline{0.911}} & 24 & \hj{1.375}\\
DAIN$_{f}$ \cite{bao2019depth} & 27.52/0.821/3.47 & 29.99/0.910 & 24 & \hj{1.375}\\
\hline
Ours (S$_{tst}$=3) & \textcolor{blue}{\underline{28.86}}/\textcolor{blue}{\underline{0.858}}/\textcolor{blue}{\underline{2.67}} & \textcolor{red}{\textbf{30.29}}/\textcolor{red}{\textbf{0.912}} & \textcolor{red}{\textbf{5.5}} & \textcolor{blue}{\underline{0.074}}\\
Ours (S$_{tst}$=5) & \textcolor{red}{\textbf{30.12}}/\textcolor{red}{\textbf{0.870}}/\textcolor{red}{\textbf{2.15}} & \textcolor{blue}{\underline{30.18}}/\textcolor{blue}{\underline{0.911}} & \textcolor{red}{\textbf{5.5}} & \hj{0.075}\\
\hline
\multicolumn{5}{l}{$\times$N: The ratio of number of iterations of the original version to that of -}\\
\multicolumn{5}{l}{ retrained version in the fair condition.\quad \#P: The number of parameters (M).}\\
\multicolumn{5}{l}{R$_{t}$: The runtime on 1024$\times$1024-sized frames in sec.}\\
\multicolumn{5}{l}{\textcolor{red}{\textbf{RED}}: Best performance, \textcolor{blue}{\underline{BLUE}}: Second best performance.}\\
\end{tabular}}
\end{center}\vspace{-2mm}
\caption{Quantitative comparisons on \jh{both X-TEST (4K) and Adobe240fps (HD)  \cite{su2017deep}} for multi-frame interpolation ($\times8$).}
\label{table:comparisons_on_X-TEST_Adobe}
\end{table}

\begin{figure*}
\centering
\includegraphics[scale=1.01]{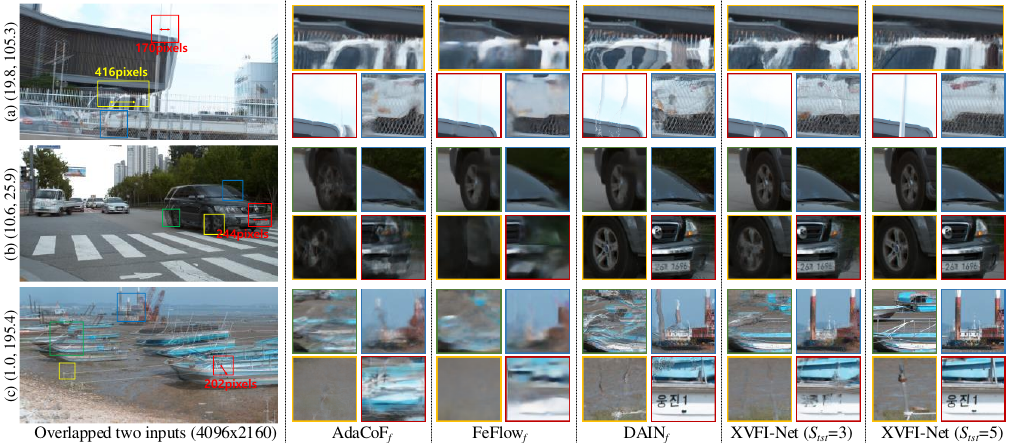}
\caption{Visual comparisons for VFI results ($t=0.5$) on X-TEST for our and retrained SOTA methods with X-TRAIN. \jh{(*,*): occlusions and optical flow magnitudes between the two input frames measured by \cite{hur2019iterative}, respectively}. \textit{Best viewed in zoom.}}
\label{fig: Visual comparison to the previous methods}
\end{figure*}

\noindent
\textbf{Quantitative Comparison.} Table \ref{table:comparisons_on_X-TEST_Adobe} shows the quantitative comparisons of the VFI methods on \jh{both X-TEST and Adobe240fps}. Please note that all runtimes (R$_{t}$) in Table \ref{table:comparisons_on_X-TEST_Adobe} are measured for 1024$\times$1024-sized frames because DAIN and FeFlow are too heavy to run for each of 4K input frames (4096$\times$2160) \textit{at once}. As shown in Table \ref{table:comparisons_on_X-TEST_Adobe}, our proposed XVFI-Net \jh{with $S_{tst}$ set to both 3 and 5 clearly outperforms} all the previous methods with large margins \jh{on both X-TEST and Adobe240fps}, even though the number of model parameters (\#P) of our model is significantly smaller than those of the others. It is also worthwhile to note that our model can infer the intermediate frames of 4K \textit{at once}, without any patch-wise iteration. In particular, XVFI-Net (S$_{tst}$=5) outperforms DAIN$_{f}$ by 2.6dB, 0.049 and 1.32 in terms of PSNR, SSIM and tOF, respectively, \jh{for X-TEST}, by utilizing only 22.9\% of DAIN's parameters. 

\jh{Especially} for the X-TEST that contains significantly extreme motions in 4K frames, our XVFI-Net can effectively capture large motion in earlier stages and then precisely interpolate the 4K input frames better than the previous methods. It is noted that FeFlow is inappropriate for large motion alignment in the feature domain, which results in blurry output and is computationally heavy for 4K input frames. In addition, the center-frame interpolation methods such as AdaCoF, FeFlow and others \cite{zhang2019flexible,gui2020featureflow,lee2020adacof,peleg2019net,niklaus2017video} tend to synthesize intermediate frames generally worse than those of arbitrary time VFI methods such as DAIN and XVFI-Net \hj{as shown in Fig. \ref{fig: multi-frame interplation}}. The errors of the center-frame interpolation methods tend to be accumulated iteratively due to inaccurate predictions. On the other hand, our model can accurately generate intermediate frames at arbitrary time \textit{t}.

\noindent
\textbf{Qualitative Comparison.} Fig. \ref{fig: Visual comparison to the previous methods} shows the visual comparison for VFI performances. The first column images in Fig. \ref{fig: Visual comparison to the previous methods} show overlapped images of two 4K input frames. As shown, huge pixel displacements are observed between two input frames, which is very challenging for VFI. The interpolated results in Fig. \ref{fig: Visual comparison to the previous methods} correspond to the center time ($t=0.5$) of the two input frames which is the most challenging frame interpolation. \jh{As shown in Fig. \ref{fig: Visual comparison to the previous methods}, our XVFI-Net ($S_{tst}=5$) surprisingly captures very complex structures of objects with extremely fast motions, which are failed by all the previous methods.}

\begin{figure}
\includegraphics[width=0.48\textwidth]{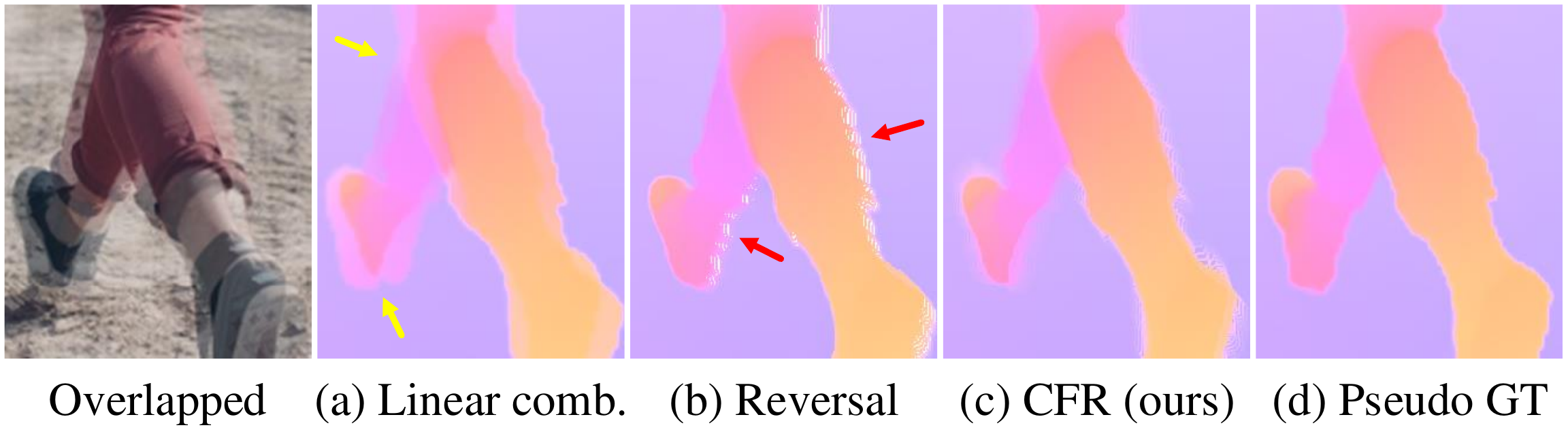}
\caption{Approximated optical flows $F_{t0}$ by three different flow approximation methods. (a) Linear combination, (b) flow reversal, (c) CFR (proposed). \textit{Best viewed in zoom.}}
\label{fig: ablation_flow}
\end{figure}


\subsection{Ablation Studies}
\noindent
\textbf{Flow Approximation.} We compare the three flow approximations that enable to produce intermediate frames at arbitrary $t$: (a) the linear approximation \cite{jiang2018super} with $F_{01}, F_{10}$, (b) flow reversal \cite{xu2019quadratic} of $F_{0t}$ and $F_{1t}$, and (c) our proposed complementary flow reversal (CFR). \hj{In this comparison, we approximated $F_{t0}$ with the three methods using the estimated optical flows $F_{01}, F_{10}$ which are obtained by IRR-PWC \cite{hur2019iterative} between the input $I_0$ and $I_1$. The importance mask z's in Eq. \ref{eq_fwarp1} and \ref{eq_fwarp2} are excluded in this comparison. Fig. \ref{fig: ablation_flow} visualizes an example of the approximated optical flows by the three methods and the pseudo ground truth which is estimated between $I_t$ and $I_0$ by IRR-PWC \cite{hur2019iterative}. To evaluate the flow approximations quantitatively, the averaged endpoint errors (EPEs) for the three methods are calculated between the approximated flows and the pseudo ground truth on the testset of Vimeo90K \cite{xue2019video}, which are shown in Table \ref{table:flow_ablation}.}
\hj{The linear approximation reveals misalignment due to the different anchor frames, which is indicated by yellow arrows in Fig. \ref{fig: ablation_flow}.} The flow reversal resolves the misalignment problem, but is inferior to the linear approximation because it causes holes that are not projected from any flow vector, as shown in the second optical flow map (red arrows). \hj{Also, the EPE value of the flow reversal is the worst among the three methods.} On the other hand, our proposed CFR can appropriately fill in the holes since the bidirectional flows complement each other, as shown in Fig. \ref{fig: ablation_flow}, \hj{which is consistent with the lowest EPE value of CFR in Table \ref{table:flow_ablation}.}

\hj{In order to investigate the efficacy of the proposed CFR for VFI, we trained three VFI models from scratch by adopting each of the three flow approximations in their BiOF-T modules, \textit{without} any help of pretrained networks.} The lowest scale depths for both training $S_{trn}$ and test $S_{tst}$ are set to 3. The VFI performances on our X-TEST (PSNR/SSIM/tOF) for the three models are listed in Table \ref{table:flow_ablation}, showing a superiority of our proposed CFR.

\begin{table}
\begin{center}
\scalebox{0.85}{
\begin{tabular}{ c||c||c|c|c }
\hline
\backslashbox{Methods}{Metrics} & EPE$\downarrow$  & PSNR$\uparrow$ & SSIM$\uparrow$ & tOF$\downarrow$ \\ \hline
(a) Linear comb. & 0.0752 & 28.73 & 0.8518 &  2.89 \\ \hline
(b) Flow reversal & 0.0892 & 28.30 & 0.8425 & 2.98 \\ \hline
(c) CFR (ours) & \textbf{0.0721} & \textbf{28.86} & \textbf{0.8582} & \textbf{2.67} \\ \hline
\end{tabular}}\vspace{-3mm}
\end{center}
\caption{\hj{The endpoint error (EPE) between the approximated $F_{t0}$ and the pseudo ground truth is obtained by IRR-PWC \cite{hur2019iterative} on Vimeo90K \cite{xue2019video} testset. Note that the VFI performances are measured on X-TEST in terms of PSNR, SSIM and tOF for three models that adopt each approximation method.}}
\label{table:flow_ablation}
\end{table}


\noindent
\textbf{Adjustable Scalability}. The lowest scale depth $S_{tst}$ for the inference can be adaptive to the degree of motion magnitudes and spatial resolution of the input frames, even after once trained, as shown in Fig. \ref{fig: framework}. We show the adjustable scalability of our framework with $S_{trn} = 1,3,5$ for $S_{tst} = 1,3,5$. For this, we train XVFI-Net variants by fully utilizing $512\times512$-sized patches because the spatial resolution of the training inputs should be multiple of 512 for $S_{trn} = 5$ where the number 512 is determined as $2^{S_{trn}=5}\times$ $M(=4)$ $\times$ 4 (via the bottlenecks of the autoencoders). Table \ref{table:Ablation study on adjustable scalability} compares the performances of the XVFI-Net variants. As shown in Table \ref{table:Ablation study on adjustable scalability}, the performances are generally boosted by increasing the value of $S_{tst}$ with the help of effectively enlarging receptive field sizes and elaborately refining the resulting flows, especially in capturing extremely large motions and detailed structures. \jh{This trend is also observed in Table \ref{table:comparisons_on_X-TEST_Adobe} for the XVFI-Net trained with $384\times384$-sized patches of $S_{trn} = 3$. \hj{Furthermore,} as shown in the rightmost four columns of Fig. \ref{fig: Visual comparison to the previous methods}, the details of the objects, letters and textures are more precisely synthesized for $S_{tst} = 5$ than 3 \hj{qualitatively}. Both quantitative and qualitative results clearly show the effectiveness of the XVFI-Net's adjustable scalability.} \hj{On the other hand, the occlusions and flow magnitudes of the Adobe240fps dataset \cite{su2017deep} are much smaller than those of X-TEST as shown in Table \ref{dataset_statistics}. It is noted in Table \ref{table:comparisons_on_X-TEST_Adobe} that our XVFI-Net with $S_{tst} = 3$ shows better performance than that with $S_{tst} = 5$ on the Adobe240fps dataset with smaller resolutions than X-TEST, which also clearly supports the efficacy of our adjustable scalability.}

\begin{table}
\begin{center}
\scalebox{0.76}{
\begin{tabular}{ c||c|c|c }
\hline
\multirow{2}{*}{\backslashbox{$S_{trn}$}{$S_{tst}$}} & \multicolumn{3}{c}{(PSNR(dB)$\uparrow$ / SSIM\cite{wang2004image}$\uparrow$ / tOF\cite{Chu2020TecoGAN}$\downarrow$)}\\
& 1 & 3 & 5 \\
\hline
1 & 26.85/0.806/4.90 & \textcolor{red}{\textbf{28.40}}/\textcolor{red}{\textbf{0.852}}/\textcolor{red}{\textbf{3.46}} & 27.14/0.842/3.69 \\ 
\hline
3 & 23.61/0.729/6.56 & 29.22/0.863/2.68 & \textcolor{red}{\textbf{30.35}}/\textcolor{red}{\textbf{0.879}}/\textcolor{red}{\textbf{1.98}} \\
\hline
5 & 22.37/0.699/6.71 & 23.70/0.724/6.39 & \textcolor{red}{\textbf{29.48}}/\textcolor{red}{\textbf{0.864}}/\textcolor{red}{\textbf{2.08}} \\
\hline
\multicolumn{4}{l}{\textcolor{red}{\textbf{RED}}: Best performance of each row}\\
\end{tabular}
}
\end{center}\vspace{-3mm}
\caption{Ablation study on adjustable scalability \hj{depending on the lowest scale depth $S_{trn}$ and $S_{tst}$ measured on X-TEST.}}
\label{table:Ablation study on adjustable scalability}
\end{table}



\begin{figure}
\includegraphics[scale=0.7]{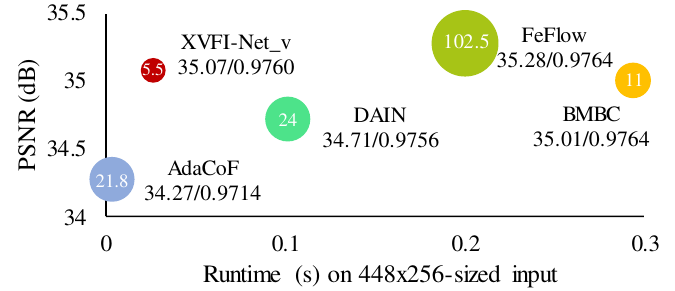}
\caption{PSNR/SSIM vs runtime (s) on Vimeo90K \cite{xue2019video} with model size (M) indicated in each circle.}
\label{fig: PSNR_vs_Runtime}
\vspace*{-0.1in}
\end{figure}

\noindent
\jh{\textbf{Robustness of our XVFI-Net Framework}}. To show the robustness of our XVFI-Net framework for LR-LFR benchmark dataset, we construct a variant of XVFI-Net, called XVFI-Net$_{v}$, \jh{with $M\!=\!2$} for the dataset with lower resolution frames. The XVFI-Net$_{v}$ is then trained on a standard VFI dataset that is the Vimeo90K \cite{xue2019video} training set with 51,313 triplets ($t\!=\!0.5$) of $448\!\times\!256$ size. The training went through 200 epochs with randomly cropped $256\!\times\!256$-sized patches and a mini-batch size of \hj{16}, where both $S_{trn}$ and $S_{tst}$ are set to 1. We compare our XVFI-Net$_{v}$ with \hj{four} SOTA methods: DAIN \cite{bao2019depth}, FeFlow \cite{gui2020featureflow}, AdaCoF \cite{lee2020adacof} and BMBC \cite{park2020bmbc}, where their pretrained models and testing code are publicly available. Fig. \ref{fig: PSNR_vs_Runtime} shows the PSNR/SSIM and runtime (s) performances of our and SOTA methods with their model sizes (M) evaluated on Vimeo90K testset. As shown, our XVFI-Net$_{v}$ outperforms BMBC, DAIN and AdaCoF with a significantly smaller model size (5.5 million parameters), by taking advantage of the recursive multi-scale and shared structure. However, the XVFI-Net$_{v}$ shows lower performance than that of FeFlow but has a much smaller model size only with $5.4\%$ of the number of the FeFlow's parameters, thus leading to \hj{about $\times 7$} faster runtime. As a result, our XVFI-Net framework designed for high-resolution VFI with extremely large motion shows its robustness to the LR-LFR benchmark dataset by simply adjusting \jh{module scale factor $M$}, $S_{trn}$ and $S_{tst}$.

\vspace{-0.5mm}
\section{Conclusion}
We first proposed a high-quality HFR dataset in HR, called X4K1000FPS with a wide range of motions. The proposed XVFI-Net can handle large pixel displacements with an adjustable scalability for inference to cope with the input resolutions or the motion magnitudes, even if training is once over. The XVFI-Net showed state-of-the-art performance on HR datasets compared to the previous methods and its robustness to the LR-LFR benchmark dataset.

Although our proposed X4K1000FPS dataset was obtained by using one single camera, such an extreme HFR 4K dataset is very valuable for the research community of VFI because such kinds of cameras are few. Besides, we delicately select clips as X-TRAIN/X-TEST to be publicly available by considering both occlusions and flow magnitudes for a new challenging VFI task, called the eXtreme Video Frame Interpolation (XVFI). We hope that this research would be a valuable milestone to extend the current VFI for more recent real-world applications with HR video.

\smallskip\noindent
\textbf{Acknowledgement.}
This work was partly supported by Institute of Information \& communications Technology Planning \& Evaluation (IITP) grant funded by the Korea government (MSIT) (No. 2017-0-00419, Intelligent High Realistic Visual Processing for Smart Broadcasting Media, 100\%).

{\small
\bibliographystyle{ieee_fullname}

}

\clearpage
\twocolumn[{%
\renewcommand\twocolumn[1][]{#1}%
\begin{center}\centering
    \setlength{\tabcolsep}{0.06cm}
    \setlength{\columnwidth}{4.cm}
    \hspace*{-\tabcolsep}\begin{tabular}{cccc}
            \animategraphics[width=\columnwidth, poster=3, autoplay, loop, final, nomouse, method=widget]{15}{VideoFigure_supp/013/}{00000}{00039}
        &
            \animategraphics[width=\columnwidth, poster=3, autoplay, loop, final, nomouse, method=widget]{15}{VideoFigure_supp/016/}{00000}{00039}
        &
            \animategraphics[width=\columnwidth, poster=3, autoplay, loop, final, nomouse, method=widget]{15}{VideoFigure_supp/027/}{00000}{00039}
        &
            \animategraphics[width=\columnwidth, poster=3, autoplay, loop, final, nomouse, method=widget]{15}{VideoFigure_supp/095/}{00000}{00039}
        \\
            \footnotesize (a) 9.9 
        &
            \footnotesize (b) 140.6 
        &
            \footnotesize (c) 122.6 
        &
            \footnotesize (e) 2.1 
        \\ 
        
            \animategraphics[width=\columnwidth, poster=3, autoplay, loop, final, nomouse, method=widget]{15}{VideoFigure_supp/104/}{00000}{00039}
        &
            \animategraphics[width=\columnwidth, poster=3, autoplay, loop, final, nomouse, method=widget]{15}{VideoFigure_supp/110/}{00000}{00039}
        &
            \animategraphics[width=\columnwidth, poster=3, autoplay, loop, final, nomouse, method=widget]{15}{VideoFigure_supp/128/}{00000}{00039}
        &
            \animategraphics[width=\columnwidth, poster=3, autoplay, loop, final, nomouse, method=widget]{15}{VideoFigure_supp/136/}{00000}{00039}
        \\
            \footnotesize (f) 12.3 
        &
            \footnotesize (g) 33.7 
        &
            \footnotesize (h) 29.7 
        &
            \footnotesize (i) 66.3 
        \\ 
        
            \animategraphics[width=\columnwidth, poster=3, autoplay, loop, final, nomouse, method=widget]{15}{VideoFigure_supp/138/}{00000}{00039}
        &
            \animategraphics[width=\columnwidth, poster=3, autoplay, loop, final, nomouse, method=widget]{15}{VideoFigure_supp/145/}{00000}{00039}
        &
            \animategraphics[width=\columnwidth, poster=3, autoplay, loop, final, nomouse, method=widget]{15}{VideoFigure_supp/152/}{00000}{00039}
        &
            \animategraphics[width=\columnwidth, poster=3, autoplay, loop, final, nomouse, method=widget]{15}{VideoFigure_supp/164/}{00000}{00039}
        \\
            \footnotesize (j) 166.8 
        &
            \footnotesize (k) 30.7 
        &
            \footnotesize (l) 47.3 
        &
            \footnotesize (m) 7.7 
        \\
    \end{tabular}
	
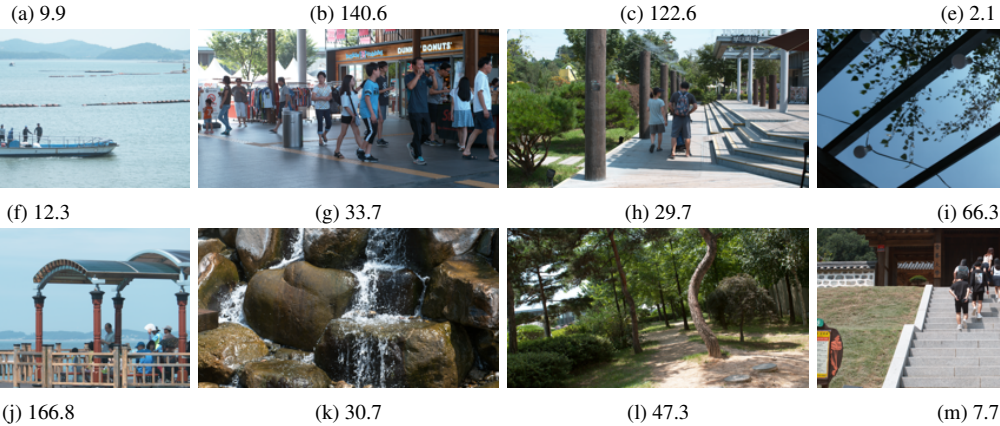
\captionof{figure}{
	More examples of our X4K1000FPS dataset, which contain diverse motions in 4K-resolution of 1000 fps. The numbers below the examples are the magnitude means of optical flows between two input frames in 30 fps. This is a video figure that can be best viewed with motion using Adobe\texttrademark\ Reader. It should be noted that they are rendered in down-scales at 15 fps for visualization convenience.
	}
	\label{fig:X4K1000FPS_suppl}
\end{center}
}]

\begin{appendices}

\section{Details of Proposed X4K1000FPS Dataset}
\subsection{Photographing Videos}
In order to provide a wide range of object motions and various camera motion types at different speeds in diverse locations, the shooting rules were guided as follows: (i) shooting various objects independently moving while the camera is stationary, (ii) shooting videos from a moving car (fast translated videos), (iii) shooting while walking (moving at normal speed), (iv) shooting with the camera in irregular motion trajectories at non-uniform speeds, (v) shooting with zooming out or in and panning at the same time. Besides, the contents of the videos also include various objects (crowds, cars, trains, plants, animals, boats, traffic signs, signboards, waterfalls, buildings, etc.) in various types of places such as stadiums, stations, beaches, markets, parks, rivers, playgrounds, etc. Fig. \ref{fig:X4K1000FPS_suppl} shows some representative thumbnails of spatiotemporally down-sampled 4K video at 15 fps for a visualization purpose. As shown in Fig. 1 in the main paper and Fig. \ref{fig:X4K1000FPS_suppl} in this Supplementary Material, very extreme scenes including various camera motions, zooming, translations, speeds, occlusions and objects are contained in the X4K1000FPS dataset. 


\subsection{Test Dataset: X-TEST}
\hj{We manually select the consecutive 32 frames for each test scene by considering the degrees of occlusion, optical flow magnitudes and diversity among 5,000 frames. We compose nonuplets by sampling every 4 frames from 32 frames of each test scene. Since the frame rates of videos are often given in multiples of 30 in VFI benchmark datasets \cite{su2017deep, xue2019video} or real-world industries, we approximate that our test videos are set to 960 fps (= 32$\times$30 fps) instead of 1,000 fps. Therefore, two input frames that are 32 frames apart are regarded as part of a 30 fps (= 960 / 32) video and converted to 240 fps by $\times8$ multi-frame interpolation in the evaluation phase.}

\subsection{Train Dataset: X-TRAIN}
\hj{To compose a valuable training dataset from our enormous videos for video frame interpolation, we select training samples based on the value of the occlusion map estimated by IRR-PWC \cite{hur2019iterative}. The occlusion maps are approximated on the spatially down-sampled ($\times1/4$) and temporally sub-sampled ($\times1/32$) frames of the original 4K-resolution 1000 fps videos. Each target frame is fed into IRR-PWC with the previous and the next frames of the target frame, respectively. The resulting two occlusion maps are averaged to get the bidirectional occlusion map.}

\hj{Then, we divide the 4K-resolution frame into overlapping patches of $768\times768$, which forms an $81\times31$ grid, except for the boundary of the 4K-resolution frame. This is because the boundary patches have undesirably large occlusion values when there are translation motions. Similarly, about 5,000 frames are divided into overlapping 154 clips of 65 consecutive frames except for the boundary period in the temporal dimension of the scenes. Thus, about 386K (= $81\times31\times154$) candidate training samples per scene of the patch size of 768$\times$768 and the lengths of 65 frames are extracted from a 4K-resolution 1000 fps video. After that, the candidate training samples whose bidirectional occlusion value is the top 10\% of those of all candidate training samples remained, and the others are discarded. Finally, total 4,408 training samples are sparsely selected as training data to prevent similar samples from being selected to maintain the diversity of the training samples.}


\section{Details of Architecture of XVFI-Net}
In addition to Fig. 3 and 4 in the main paper, we present the detailed architectures of sub-networks of XVFI-Net \hj{in the case of the module scale factor $M=4$} in Table \ref{tab:FeatExtraction} to Table \ref{tab:Refinement}. The series of rows represents the consecutive operations. The first column represents each layer's operation, and H,W and C indicate the spatial ratio with respect to bicubicly downsampled ($\times1/2^s$) input frames $I^s_i$ for each scale $s$ and the number of channels of the output tensors, respectively. The last column denotes the names of some output tensors, which are worth mentioning. We omit the names of the output tensors if they are just intermediate tensors in the sub-networks. When the multiple tensors are input to each layer, they are concatenated channel-wise. `resblock' represents a residual block which consists of \textit{conv2d} - \textit{relu} - \textit{conv2d} - \textit{identity addition}. The stride of the convolutional layer is set to 1, if not mentioned. The convolution filter sizes are $3\times3$ and $4\times4$ for the strides of 1 and 2, respectively. 

As shown in Table \ref{tab:FeatExtraction}, the Feature Extraction Block is a simple residual block-based sub-network. On the other hand, the flow estimation sub-networks, the BiFlownet and TFlownet, have a simple auto-encoder architecture to enlarge the receptive field as shown in Table \ref{tab:BiFlownet_lowest}, \ref{tab:BiFlownet} and \ref{tab:TFlownet}. \hj{The Refinement Block has a U-Net \cite{ronneberger2015u}-based architecture as in Table \ref{tab:Refinement}.} The parameters of each sub-network are shared for all scale levels except for the BiFlownet at the lowest scale depth $S$, which is isolated in Table \ref{tab:BiFlownet_lowest}. \hj{The bidirectional flows are estimated directly from two input features $C^S_0, C^S_1$ at the lowest scale level $S$, because there does not exist any provided initial flow.}

\textbf{Efficiency of XVFI-Net During Inference.}\quad \hj{The BiOF-T module can start from any down-scaled level, while the BiOF-T module can be skipped in the down-scaled levels ($s=1,\dots,S_{tst}$) as described in Fig. 3 in the main paper. By doing so, our XVFI-Net framework can accelerate run time \hj{about $22\%$} faster compared to the full-recursion framework where both BiOF-I and -T modules are processed together in all scale levels for 4K video, when $S_{tst}=5$.} Besides, the additional runtimes induced by the smaller down-scaled levels ($s>0$) are negligible since the runtimes of the BiOF-I module at down-scaled levels are much smaller than those of BiOF-I and BiOF-T modules at the original scale ($s=0$).

\begin{table*}
    \begin{center}
    \begin{tabular}{|c|c|c|c|}
    \hline
    Operation & H,W & C & Remarks \\
    \hline
    input $I^0_i$ & 1 & 3 & $I^0_i$ \\
    conv2d - relu & 1 & 64 & - \\
    conv2d - relu & 1/2 & 64 & - \\
    conv2d & 1/4 & 64 & $\text{Feat}_i$ \\
    resblock ($\times2$) - add to $\text{Feat}_i$ & 1/4 & 64 & $C^0_i$ \\
    \hline
    conv2d ($\times s$) & 1/4 $\times$ $1/2^s$  & 64 & $C^s_i$ \\
    \hline
    \end{tabular}
    \end{center}
    \caption{The detailed architecture of the Feature Extraction Block of XVFI-Net. $C^s_i$ is obtained by applying the last convolutional layer to $C^0_i$ $s$ times recurrently. The parameters are temporally shared for the two input frames ($i=0, 1$).}
\label{tab:FeatExtraction}
\end{table*}

\begin{table*}
    \begin{center}
    \begin{tabular}{|c|c|c|c|}
    \hline
    Operation & H,W ($\times$ $1/2^S$) & C & Remarks \\
    \hline
    input [$C^S_0$, $C^S_1$] & 1/4 & 64$\times$2 & - \\
    conv2d (stride 2) - relu & 1/8 & 128 & - \\
    conv2d (stride 2) - relu & 1/16 & 256 & - \\
    NN upscale - conv2d - relu & 1/8 & 128 & - \\
    NN upscale - conv2d - relu & 1/4 & 64 & - \\
    conv2d & 1/4 & 2+2+1+1 & $F^S_{01}$, $F^S_{10}$, $z^S_{01}$, $z^S_{10}$ \\
    \hline
    \end{tabular}
    \end{center}
    \caption{The detailed architecture of the auto-encoder-based BiFlownet of XVFI-Net at the lowest scale depth.}
\label{tab:BiFlownet_lowest}
\end{table*}

\begin{table*}
    \begin{center}
    \begin{tabular}{|c|c|c|c|}
    \hline
    Operation & H,W ($\times$ $1/2^s$) & C & Remarks \\
    \hline
    input [$C^s_0$, $\Tilde{C}^s_{01}$] & 1/4 & 64$\times$2 & - \\
    conv2d & 1/4 & 64 & $\hat{C}^s_{01}$ \\
    \hline
    input [$C^s_1$, $\Tilde{C}^s_{10}$] & 1/4 & 64$\times$2 & - \\
    conv2d & 1/4 & 64 & $\hat{C}^s_{10}$ \\
    \hline
    input [$\hat{C}^s_{01}$, $\hat{C}^s_{10}$,$\Tilde{F}^s_{01}$, $\Tilde{F}^s_{10}$] & 1/4 & 64$\times$2+2$\times$2 & - \\
    conv2d (stride 2) - relu & 1/8 & 128 & - \\
    conv2d (stride 2) - relu & 1/16 & 256 & - \\
    NN upscale - conv2d - relu & 1/8 & 128 & - \\
    NN upscale - conv2d - relu & 1/4 & 64 & - \\
    conv2d - add to [$\Tilde{F}^s_{01}$, $\Tilde{F}^s_{10}$] & 1/4 & 2+2+1+1 & $F^s_{01}$, $F^s_{10}$, $z^s_{01}$, $z^s_{10}$ \\
    \hline
    \end{tabular}
    \end{center}
    \caption{The detailed architecture of the auto-encoder-based BiFlownet of XVFI-Net except for the lowest scale depth.}
\label{tab:BiFlownet}
\end{table*}

\begin{table*}
    \begin{center}
    \begin{tabular}{|c|c|c|c|}
    \hline
    Operation & H,W ($\times$ $1/2^s$) & C & Remarks \\
    \hline
    input [$C^s_0$, $C^s_1$, $\Tilde{C}^s_{t0}$, $\Tilde{C}^s_{t1}$,$\Tilde{F}^s_{t0}$, $\Tilde{F}^s_{t1}$] & 1/4 & 64$\times$4+2$\times$2 & - \\
    conv2d (filter 1$\times$1) - relu & 1/4 & 64 & - \\
     conv2d (stride 2) - relu & 1/8 & 128 & - \\
     conv2d (stride 2) - relu & 1/16 & 256 & - \\
     NN upscale - conv2d - relu & 1/8 & 128 & - \\
     NN upscale - conv2d - relu & 1/4 & 64 & - \\
     conv2d - add to [$\Tilde{F}^s_{t0}$, $\Tilde{F}^s_{t1}$] & 1/4 & 2+2+1 & $F^s_{t0}$, $F^s_{t1}$, $m^s$ \\
    \hline
    \end{tabular}
    \end{center}
    \caption{The detailed architecture of the auto-encoder-based TFlownet of XVFI-Net.}
\label{tab:TFlownet}
\end{table*}

\begin{table*}
    \begin{center}
    \begin{tabular}{|c|c|c|c|}
    \hline
    Operation & H,W ($\times$ $1/2^s$) & C & Remarks \\
    \hline
    input ([${C}^s_0$, ${C}^s_1$, $\Tilde{C}^s_{t0}$, $\Tilde{C}^s_{t1}$]) & 1/4 & 256 & - \\
    pixel-shuffle \cite{shi2016real} ($\uparrow$4) & 1 & 16 & \text{PS} \\
    \hline
    input [\text{PS}, ${F}^s_{t0}\uparrow_2$, ${F}^s_{t1}\uparrow_2$, ${I}^s_0$, ${I}^s_1$, $\Tilde{I}^s_{t0}$, $\Tilde{I}^s_{t1}$] & 1 & 16+2$\times$2+3$\times$4 & - \\
    conv2d (stride 2) - relu & 1/2 & 64 & $\text{enc}_1$ \\
    conv2d (stride 2) - relu & 1/4 & 128 & $\text{enc}_2$ \\
    conv2d (stride 2) - relu & 1/8 & 256 & - \\
    conv2d - relu & 1/8 & 256 & - \\
    NN upscale - concat to $\text{enc}_2$ & 1/4 & 384 & - \\
    conv2d - relu & 1/4 & 128 & - \\
    NN upscale - concat to $\text{enc}_1$ & 1/2 & 192 & - \\
    conv2d - relu & 1/2 & 64 & - \\
    NN upscale - conv2d & 1 & 1+3 & $m^s$, $\Tilde{I}^s_{r}$ \\
    \hline
    \end{tabular}
    \end{center}
    \caption{The detailed architecture of the U-Net \cite{ronneberger2015u}-based Refinement Block of XVFI-Net.}
\label{tab:Refinement}
\end{table*}




\begin{figure*}
\centering
\includegraphics[scale=1.15]{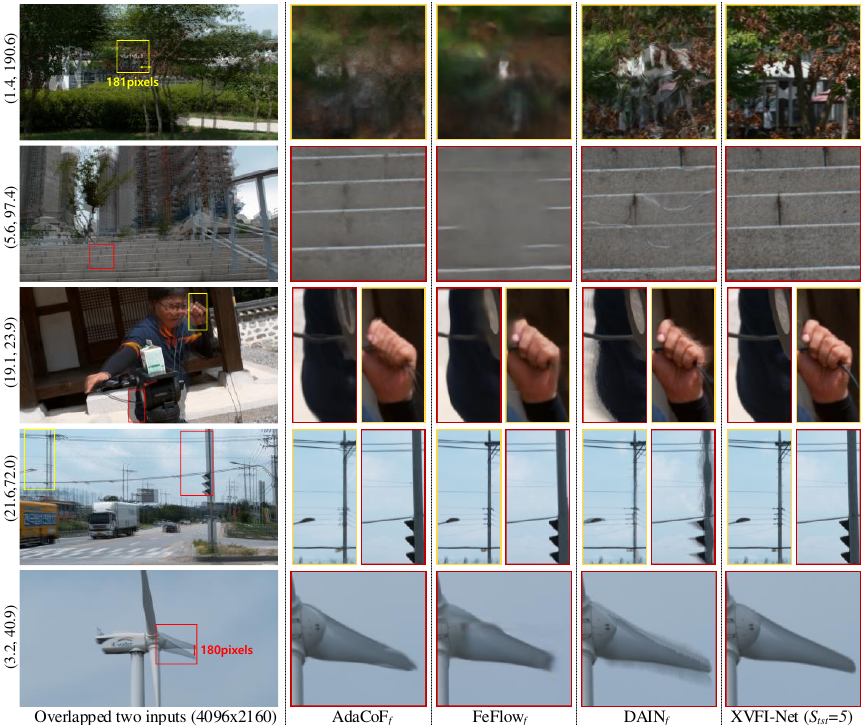}
\caption{Visual comparisons for VFI results ($t=0.5$) on X-TEST for our and \textit{retrained} SOTA methods with X-TRAIN. (*,*): occlusions and optical flow magnitudes between the two input frames measured by IRR-PWC \cite{hur2019iterative}, respectively. \textit{Best viewed in zoom.}}
\label{fig: Visual comparison to the previous methods on X-TEST}
\end{figure*}

\begin{figure*}
\centering
\includegraphics[scale=1.23]{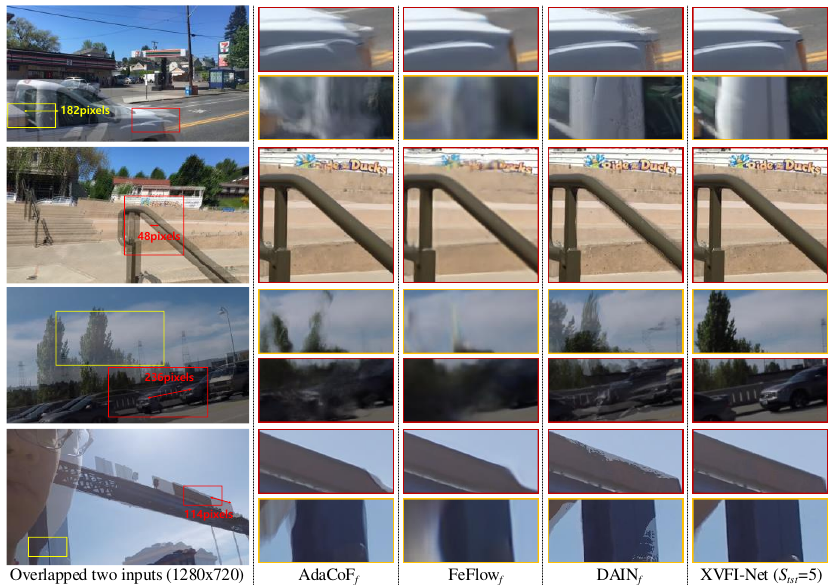}
\caption{Visual comparisons for VFI results ($t=0.5$) on Adobe240fps for our and \textit{retrained} SOTA methods with X-TRAIN. \textit{Best viewed in zoom.}}
\label{fig: Visual comparison to the previous methods on Adobe240fps}
\end{figure*}

As shown in the first scene with even extreme back and forth motions of a propeller with zoom-out, our XVFI-Net can surprisingly capture such a complex motion for VFI, but the SOTAs fail to interpolate the sophisticated tips of the propeller pointed by the yellow arrows. For the second scene, even while riding a fast-moving car, XVFI-Net better captures far tiny structures such as electric wires seen at the left part and a closer pole with a large pixel displacement pointed by the yellow arrows. For the third scene, the rightmost front car moves very fast, so all the previous methods fail to capture it, denoted by the yellow arrow, yielding severe artifacts (structural distortions). On the other hand, XVFI-Net precisely captures the especially right edges of the rightmost car. Finally, in the last scene even with the \textit{extremely} hand-shaken frames, the XVFI-Net can also synthesize repeating similar stairs but all SOTAs tend to generate \hj{baggy} artifacts. As a result, XVFI-Net significantly better handles large pixel displacements due to extreme motion and huge spatial resolutions.

\section{Additional Qualitative Results}
\begin{figure*}
\centering
\includegraphics[width=\textwidth]{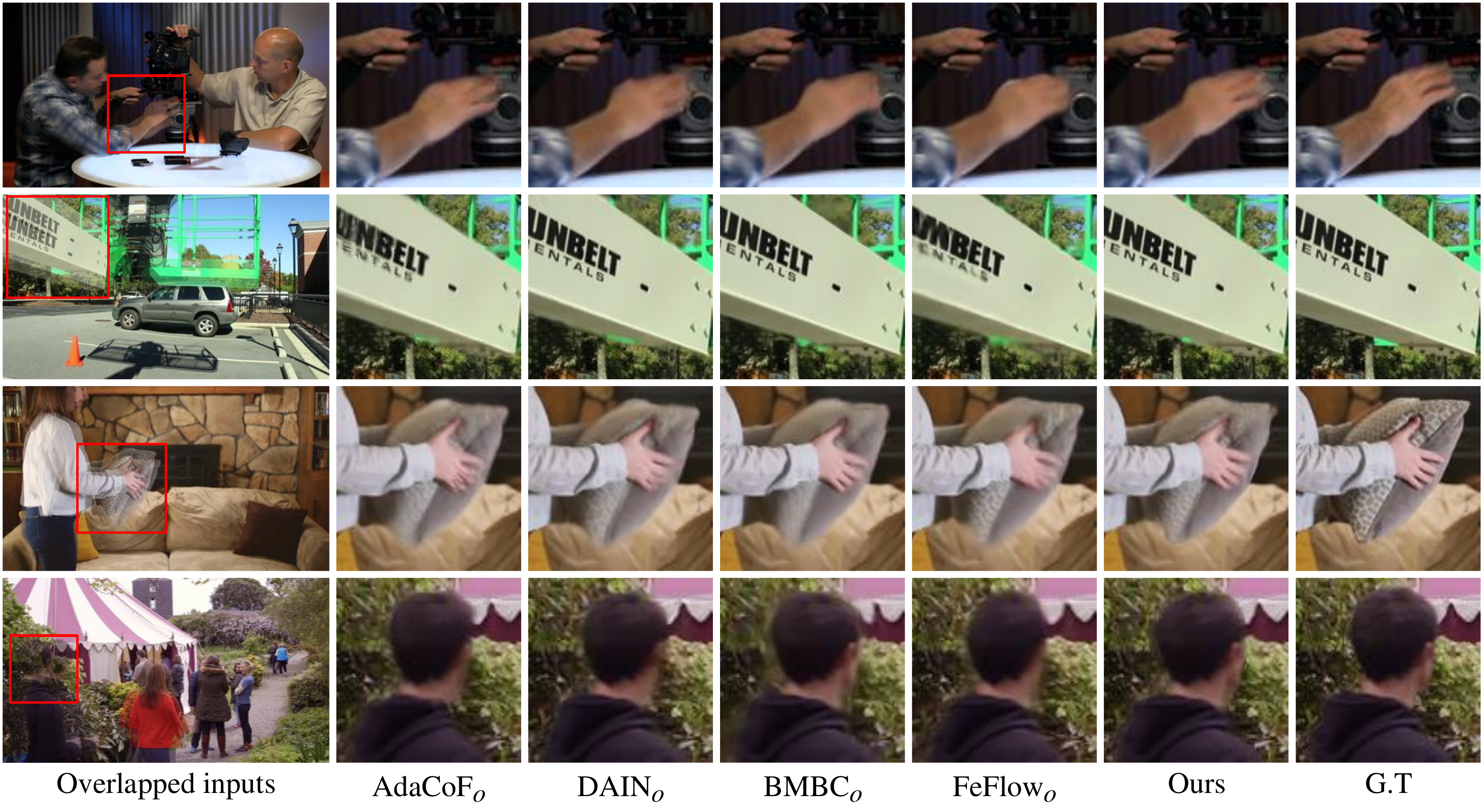}
\caption{\hj{Visual comparisons of AdaCoF \cite{lee2020adacof}, DAIN \cite{bao2019depth}, BMBC \cite{park2020bmbc}, FeFlow \cite{gui2020featureflow}, our XVFI-Net$_{v}$ and the corresponding ground truth on the testset of Vimeo90K \cite{xue2019video} triplet. \textit{Best viewed in zoom.} }}
\label{fig:vimeo_comparison}
\end{figure*}

\textbf{Visual Comparisons for VFI methods.}\quad We provide additional qualitative results on X-TEST (4K) in Fig. \ref{fig: Visual comparison to the previous methods on X-TEST}, Adobe240fps \cite{su2017deep} (HD) in Fig. \ref{fig: Visual comparison to the previous methods on Adobe240fps} and Vimeo90K \cite{xue2019video} in Fig. \ref{fig:vimeo_comparison} by each setting described in the main paper. 

\begin{figure*}
\centering
\includegraphics[width=\textwidth]{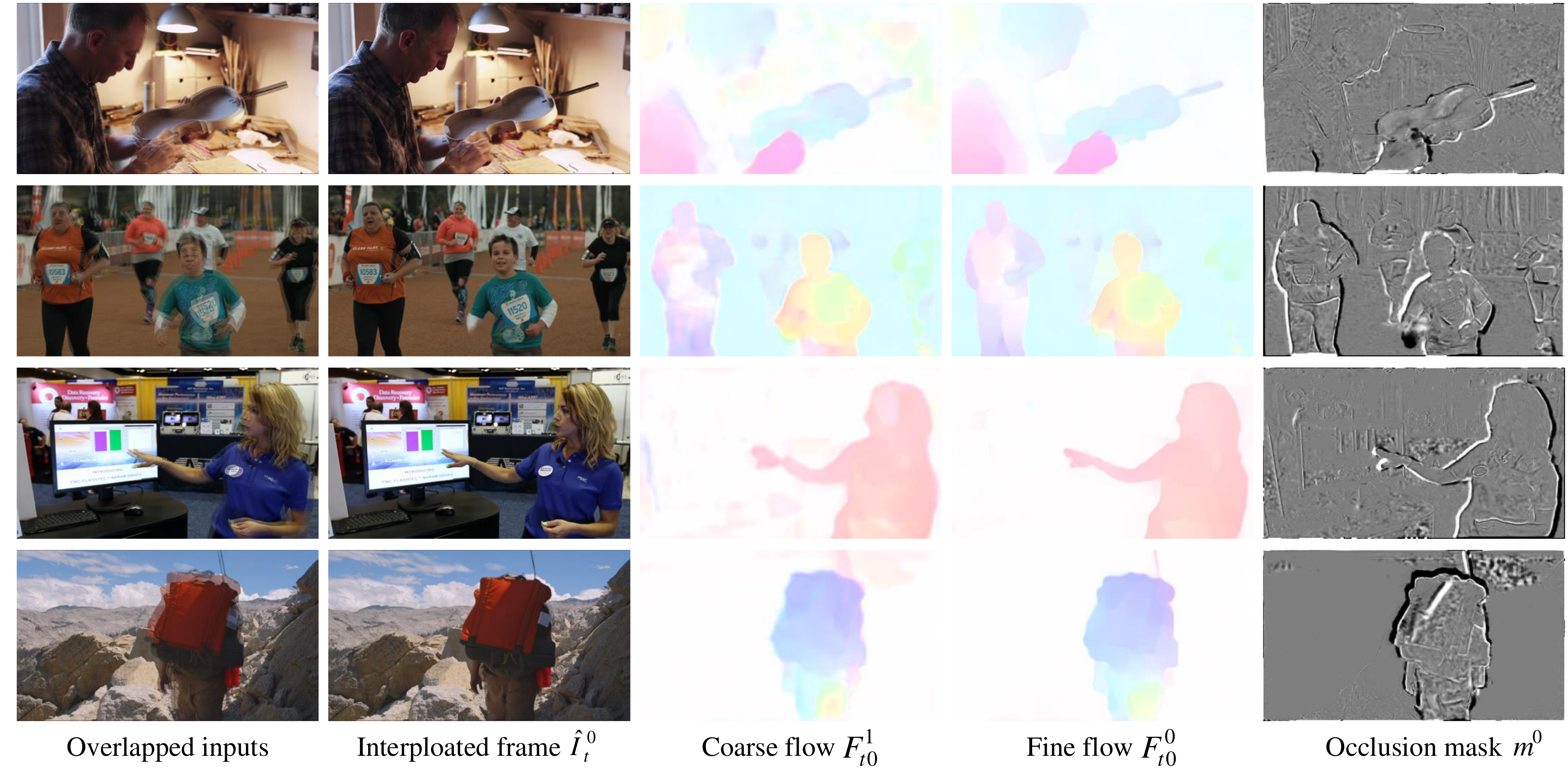}
\caption{\hj{Visualization of optical flows and occlusion masks of XVFI-Net$_{v}$. The coarse and fine flows are extracted at scale $1$ and $0$, respectively.}}
\label{fig:flow_visualize}
\end{figure*}

\textbf{Visualization of Components of XVFI-Net.}\quad Fig. \ref{fig:flow_visualize} shows the visualization of optical flows and occlusion masks of XVFI-Net$_{v}$. As expected, estimated flows at the upper level seem finer than those at the lower level ($F^{1}_{t0}$) as shown in Fig. \ref{fig:flow_visualize}. The coarse-to-fine structure gradually helps the whole XVFI framework boost the final VFI performance at original scale $s=0$ based on the occlusion masks and the iteratively updated flows that are all \textit{learned from scratch}.

\section{Failure Cases}
\begin{figure*}
\centering
\includegraphics[width=\textwidth]{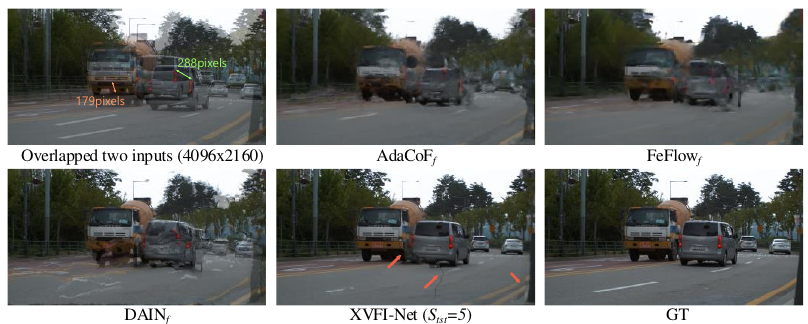}
\caption{\hj{Failure cases of 4K result ($t=0.5$) on X-TEST for our and \textit{retrained} SOTA methods with X-TRAIN, including the corresponding ground truth. \textit{Best viewed in zoom.} }}
\label{fig:Failure_Cases_4K}
\end{figure*}

\begin{figure*}
\centering
\includegraphics[width=\textwidth]{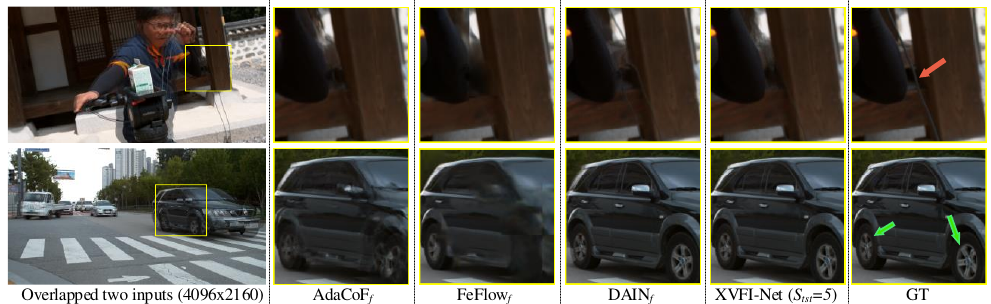}
\caption{\hj{Failure cases of cropped results ($t=0.5$) on X-TEST for our and \textit{retrained} SOTA methods with X-TRAIN, including the corresponding ground truth. \textit{Best viewed in zoom.} }}
\label{fig:Failure_Cases_Crop}
\end{figure*}

Since we delicately select the scenes to compose extremely challenging X-TEST, there exist inevitably failure patches \textit{within the same 4K frame result}, where all compared methods including XVFI-Net (S$_{tst}=5$), fail to accurately interpolate the intermediate frames. Fig. \ref{fig:Failure_Cases_4K} shows the 4K failure results ($t=0.5$) including several failure patches of ours because the input videos have very large magnitude means of optical flows (196.5) attributed to large camera shaking with the fast moving cars. Fig. \ref{fig:Failure_Cases_Crop} shows the failure cropped patches. First, the tiny electric line, which is hard to be distinguishable from static background, is failed to be accurately interpolated by all methods including ours, as indicated by a red arrow. Second, rotations of fast moving car's wheels are also challenging to be delicately synthesized while considering the degree of rotations, as pointed by two green arrows. On top of these, blurriness and abrupt brightness or color change in the input frames would also make all VFI methods challenging.

Please note that we have also provided all interpolated results for all compared methods of both original and re-trained versions on X-TEST to be publicly available at \url{https://github.com/JihyongOh/XVFI} for easier comparison.

\smallskip\noindent
\textbf{Acknowledgement.} \quad
We specially thank Sung-Jun Yoon and Hyun-Ho Kim for photographing 4K videos on the spot.

\end{appendices}

\end{document}